\documentclass{article}

\usepackage{microtype}
\usepackage{graphicx}
\usepackage{booktabs} %

\usepackage{color}
\usepackage{xcolor}
\usepackage{url}            %
\usepackage{graphicx}
\usepackage{bbm}

\usepackage{amsfonts}       %
\usepackage{nicefrac}       %
\usepackage{microtype}      %
\usepackage{amsmath, amsthm}
\usepackage{algorithmic}
\usepackage{algorithm}
\usepackage{caption}
\usepackage{subcaption}
\usepackage{wrapfig}
\usepackage[capitalise,nameinlink,noabbrev]{cleveref}

\newtheorem{definition}{Definition}

\newcommand{\task}{\mathcal{T}}

\newcommand{\state}{\mathbf{s}}

\usepackage[accepted]{icml2021}

\icmltitlerunning{CARE}

\begin{document}
\captionsetup[figure]{font=small}

\twocolumn[
\icmltitle{Multi-Task Reinforcement Learning with Context-based Representations}

\icmlsetsymbol{equal}{*}

\begin{icmlauthorlist}
\icmlauthor{Shagun Sodhani}{fb}
\icmlauthor{Amy Zhang}{fb,mila,mcgill}
\icmlauthor{Joelle Pineau}{fb,mila,mcgill}
\end{icmlauthorlist}

\icmlaffiliation{fb}{Facebook AI Research}
\icmlaffiliation{mila}{Mila}
\icmlaffiliation{mcgill}{McGill University}

\icmlcorrespondingauthor{Shagun Sodhani}{sodhani@fb.com}

\icmlkeywords{Machine Learning, ICML}

\vskip 0.3in
]

\printAffiliationsAndNotice{}

\begin{abstract}
The benefit of multi-task learning over single-task learning relies on the ability to use relations across tasks to improve performance on any single task. While sharing representations is an important mechanism to share information across tasks, its success depends on how well the structure underlying the tasks is captured. In some real-world situations, we have access to \textit{metadata}, or additional information about a task, that may not provide any new insight in the context of a single task setup alone but inform relations across multiple tasks. While this metadata can be useful for improving multi-task learning performance, effectively incorporating it can be an additional challenge.  We posit that an efficient approach to knowledge transfer is through the use of multiple context-dependent, composable representations shared across a family of tasks. In this framework, metadata can help to learn interpretable representations and provide the context to inform which representations to compose and how to compose them. We use the proposed approach to obtain state-of-the-art results in Meta-World, a challenging multi-task benchmark consisting of 50 distinct robotic manipulation tasks.

\end{abstract}

\section{Introduction}
Reinforcement learning (RL) has made large strides over the last several years~\citep{mnih2013atari,silver2017alphaGo,radford2019language_models_are_unsupervised_multitask_learners}. While these improvements are significant, much of this success has been restricted to the single task setting ~\citep{distral, meta-world}. In contrast, humans are adept at multi-tasking by acquiring new skills and composing known skills to solve complex tasks~\citep{learning_independent_causal_mechanisms, rapid_trial_and_error_learning_with_simulation_supports_flexible_tool_use_and_physical_reasoning}. For autonomous agents to adapt effectively in the real world, they need to master multiple tasks in a sample efficient manner. Multi-task reinforcement learning (MTRL) is a promising approach to train effective real-world agents~\citep{multitask_reinforcement_learning_on_the_distribution_of_mdps, policy_distillation, borsa2016mtrl, epopt_learning_robust_neural_network_policies_using_model_ensembles, scalable_multitask_policy_gradient_reinforcement_learning, modular_multitask_reinforcement_learning_with_policy_sketches, multitask_soft_option_learning, sharing_knowledge_in_multitask_deep_reinforcement_learning, gradient_surgery_for_multitask_learning}.

One limitation of existing MTRL methods is the inability to leverage side information (or \textit{metadata}), like the description of a task, to learn generalizable skills and transfer common knowledge across tasks. Such metadata is often available in real-world tasks but is not leveraged in the typical MTRL setting. This metadata can take the form of natural language task descriptions or instructions, which are often incorporated only for human usage to communicate information about tasks. These natural language descriptions have been utilized in single-task RL setups like goal-oriented RL~\citep{babyai, a_survey_of_reinforcement_learning_informed_by_natural_language, jiang2019language}. 
We show that this information can also be used to learn context-dependent, composable representations shared across a family of tasks, with the metadata acting as the context. Specifically, we show that the metadata can be used to learn a prior over a collection of encoders and can be leveraged to select the encoder(s) for any given task. The learned encoders can specialize to different aspects of the tasks, allowing for more efficient sharing of knowledge across the tasks. An important additional benefit is that the learned representations are interpretable.

\begin{figure}[b]
    \centering
    \includegraphics[height=3cm]{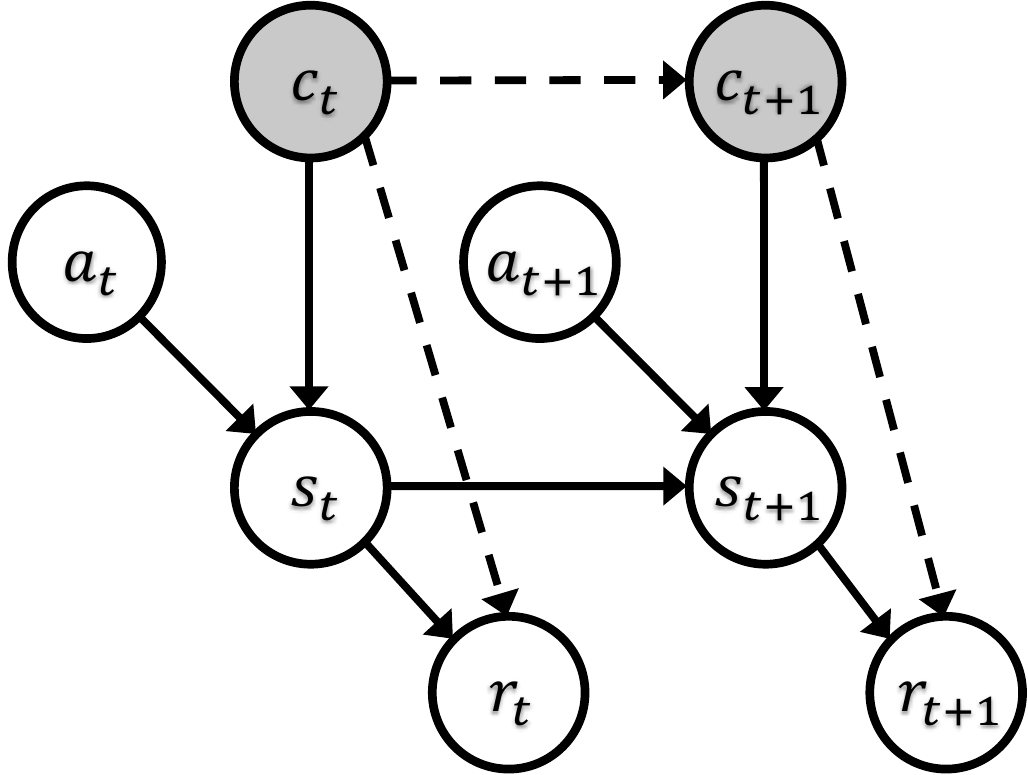} \hspace{20pt}
    \includegraphics[height=4cm]{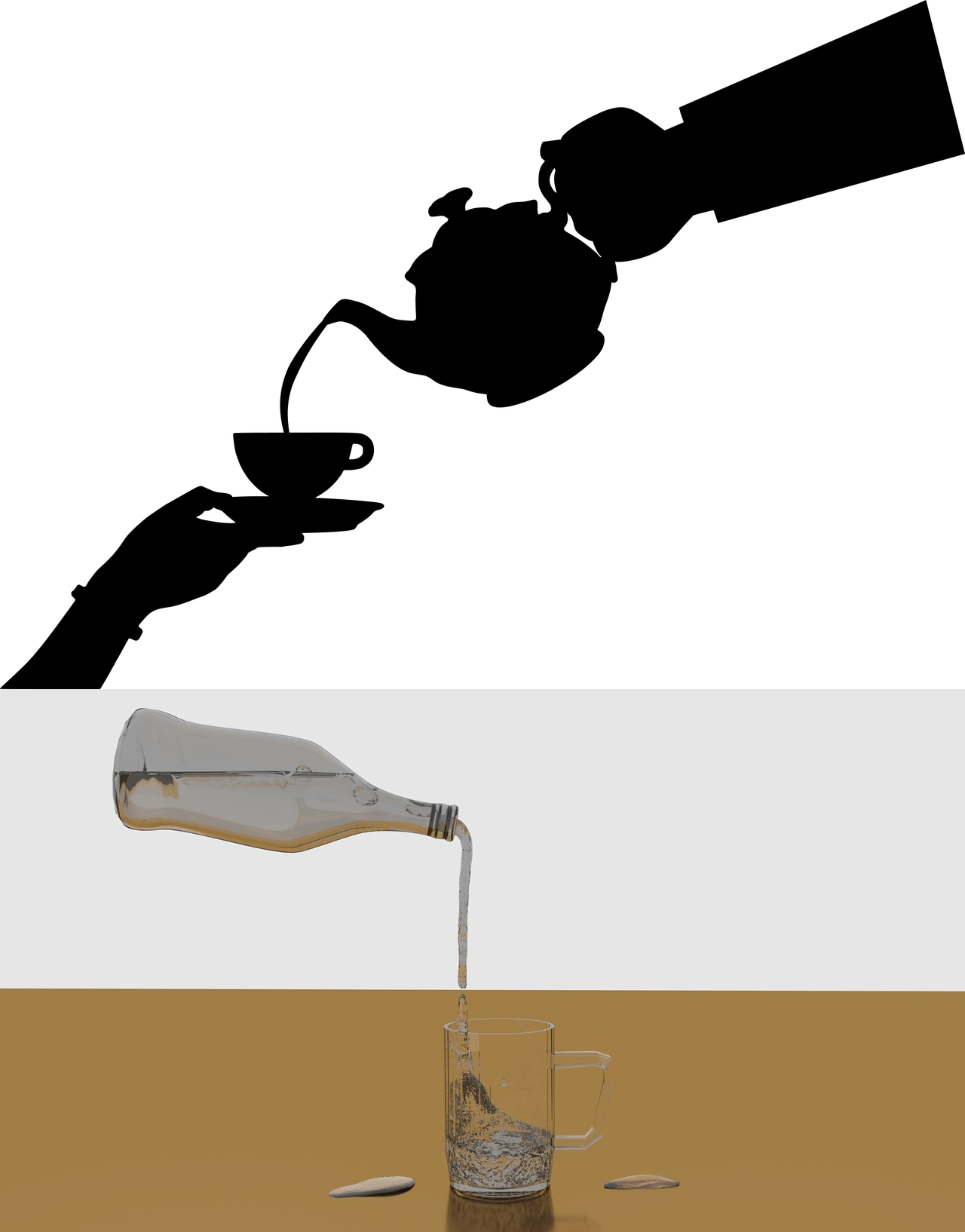}
    \caption{Graphical model of the Contextual MDP setting (left). Example of different contexts that can affect the task of pouring (right). }
    \vspace{-10pt}
    \label{fig:teaser}
\end{figure}

The default formulation of MTRL environments is as a family of Markov Decision Processes (MDPs)~\citep{Bellman1957,puterman1995markov}. This framework does not provide a natural way to incorporate metadata. We therefore turn to the \textit{Contextual Markov Decision Process} formulation~\citep{hallak2015contextual} to define our setting (\cref{fig:teaser}). We assume access to a \textit{context} (the metadata) that contains additional task-specific information which is easily available and can be used to improve per-task performance. This metadata is more useful in a multi-task setting where it can be used to infer the relationship between different tasks coming from a specified family. As an example, given two contexts, ``pour tea into my cup'' and ``pour water into the mug kept on the table'', the agent can infer the relations between the two tasks and identify the common contexts (and subtasks) like ``pour'' (\cref{fig:teaser}, right). Such context is less likely to be useful to accelerate learning in the single task setup.

Sharing representations across tasks can be an effective way for RL agents to transfer knowledge across tasks. However, not all knowledge transfers are \textit{positive}. Some aspects of a given task could be meaningful only for that task and irrelevant (or even detrimental) for the other tasks. This effect is commonly known as \textit{negative interference}~\citep{actor_mimic_deep_multitask_and_transfer_reinforcement_learning, distral}. In general, choosing which information/knowledge to transfer across the tasks, or even deciding which tasks should be learned together is an open problem~\citep{which_tasks_should_be_learned_together_in_multitask_learning?}. We posit that in the CMDP setup, the context can be used to inform which representations and information should be shared across the tasks.

In this work, we propose a novel approach for the contextual multi-task RL setting where we encode an input observation into multiple representations (corresponding to different skills or objects) using a \textit{mixture of encoders}. The learning agent can use the context to decide which representation(s) it uses for any given task, giving the agent a fine-grained control over what information is shared across tasks, thus alleviating \textit{negative interference}. We call our method \textbf{C}ontextual \textbf{A}ttention-based \textbf{RE}presentation learning, or CARE for short.

\textbf{Key contributions} of this work are 1) a simple, yet effective, way to incorporate task metadata, or contextual information, to improve sample efficiency and asymptotic performance, 2) a new representation learning algorithm for MTRL that leverages a mixture of interpretable encoders which encodes task and object-specific information about each state space, and 3) state-of-the-art results on a challenging multi-task RL benchmark, Meta-World~\citep{meta-world}. For example videos see \mbox{\url{https://sites.google.com/view/mtrl-care}}. The implementation of the algorithms is available at \mbox{\url{https://github.com/facebookresearch/mtrl}}.

\section{Preliminaries}
\label{sec:preliminaries}

A \textbf{Markov Decision Process} (MDP)~\citep{Bellman1957,puterman1995markov} is defined by a tuple $\langle {\cal S}, {\cal A}, R, T, \gamma \rangle $, where ${\cal S}$ is the set of states, ${\cal A}$ is the  set of actions, $R: {\cal S} \times {\cal A}\rightarrow \mathbb{R}$ is the reward function, $T:{\cal S} \times {\cal A} \rightarrow Dist({\cal S})$ is the environment transition probability function, and $\gamma \in [0,1)$ is the discount factor. At each time step, the learning agent perceives a state $s_t \in {\cal S}$, takes an action $a_t \in {\cal A}$ drawn from a policy $\pi : {\cal S} \times {\cal A} \rightarrow [0,1]$, and with probability $T(s_{t+1}|s_t,a_t)$ enters next state $s_{t+1}$, receiving a numerical reward $R_{t+1}$ from the environment. The value function of policy $\pi$ is defined as: $V_\pi(s) = E_\pi[\sum_{t=0}^{\infty} \gamma^{t} R_{t+1} | S_0 = s]$. The optimal value function $V^{*}$ is the maximum value function over the class of stationary policies.

\textbf{Contextual Markov Decision Processes} were first proposed by \citet{hallak2015contextual} as an augmented form of Markov Decision Processes that utilize \textit{side information} as a form of context, similar to in contextual bandits. 

\begin{definition}[Contextual Markov Decision Process]
A contextual Markov decision process (CMDP) is defined by tuple $\langle \mathcal{C}, \mathcal{S}, \mathcal{A}, \mathcal{M}\rangle$ where $\mathcal{C}$ is the context space, $\mathcal{S}$ is the state space, $\mathcal{A}$ is the action space. $\mathcal{M}$ is a function which maps a context $c\in\mathcal{C}$ to MDP parameters $\mathcal{M}(c)=\{R^c, T^c\}$.
\end{definition}

Contexts can be applied in the multi-task setting, where we define a \textit{family} of MDPs where each MDP has a shared state space $\mathcal{S}$. However, the agent only has access to a partial state space $\mathcal{S}^c$ (either low-dimensional or rich, like pixels) that is a subspace of the original state space $\mathcal{S}$, focusing only on objects relevant to the task at hand. Different MDPs can involve different combinations of objects and skills, hence the state space $\mathcal{S}^c$ and reward function $R^c$ can differ across MDPs. However, the objects are \textit{shared} across tasks, i.e., the object-specific dynamics remain consistent across tasks\footnote{Note that the dynamics $T^c$ for each MDP are still different because the state spaces are different.}. In this work, we focus on the low-dimensional setting where $\mathcal{S}^c$ is a strict subset of the dimensions in $\mathcal{S}$. 

We attach this additional relaxation of the original CMDP definition to define a new setting, a Block Contextual MDP (BC-MDP)\footnote{This is not the partial observability setting because we have access to a task id or description that uniquely identifies the task, and therefore what objects are referred to by the task-specific state space.}~\citep{du2019pcid,zhang2020invariant}: 
\begin{definition}[Block Contextual Markov Decision Process]
A block contextual Markov decision process (BC-MDP) is defined by tuple $\langle \mathcal{C}, \mathcal{S}, \mathcal{A}, \mathcal{M}'\rangle$ where $\mathcal{C}$ is the context space, $\mathcal{S}$ is the state space, $\mathcal{A}$ is the action space. $\mathcal{M}'$ is a function which maps a context $c\in\mathcal{C}$ to MDP parameters and observation space $\mathcal{M}(c)=\{R^c, T^c, \mathcal{S}^c\}$.
\end{definition}

\begin{figure}[b]
    \centering
    \includegraphics[width=0.9\linewidth]{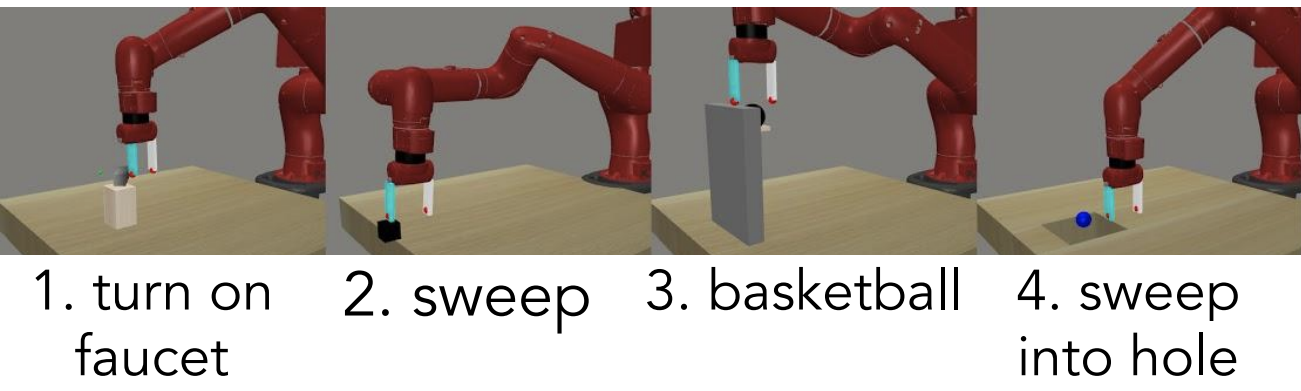}
    \vspace{-10pt}
    \caption{Some tasks from Meta-World. Image taken from~\citet{meta-world}.}
    \label{fig:metaworld}
\end{figure}

This brings us to our setting for evaluation, \textbf{Meta-World}~\citep{meta-world}\footnote{We use the following commit from MetaWorld for our experiments: af8417bfc82a3e249b4b02156518d775f29eb289}, as a natural instantiation of a BC-MDP. Meta-World proposes a benchmark for meta-RL and multi-task RL, consisting of 50 distinct robotics manipulation tasks, with some example tasks shown in \cref{fig:metaworld}. The state space across all tasks is of the same dimensionality, but those dimensions have different semantics across tasks. For example, the same subset of dimensions can refer to a goal position in one task and some object's position in the other task. We assume access to a family of $N$ MDPs consisting of potentially different reward functions and state spaces of consistent dimensionality, but not necessarily with the same semantic meaning. Unlike previous works, which generally focus on very narrow task distributions, Meta-World provides a diverse task distribution with 50 different tasks involving objects like doors, cups, windows, drawers, etc. and skills like push, pull, open, close, etc. while still providing a shared state and action space. Evaluating on a broad task distribution provides a better estimate of the generalization capabilities of MTRL algorithms.

\section{A Method for Learning Contextual Attention-based Representations}
In the multi-task reinforcement learning setting described in \cref{sec:preliminaries}, we propose to factorize the state representation into sub-components that are common across the MDPs within a defined BC-MDP family.
While each task has its own state space, there are commonalities across tasks. For example, this can take the form of objects like ``drawer'' or ``door'', and skills like ``open'' or ``close''.  In this example, the goal would be to disentangle the state into object-specific and skill-specific representations. We train a universal policy (i.e. one shared policy for all the tasks) that uses the task-specific metadata (or context) to choose a functional representation (e.g. of objects and skills) for any given task. We introduce this compositionality by incorporating a mixture of encoders where different encoders specialize to different aspects for the given family of tasks. In our example, given three tasks ``open a door'', ``open a drawer'', and ``close a drawer'', the encoders could specialize to ``open'' and ``close'' skills and ``door'' and ``drawer'' objects.

In this section, we describe how we use the metadata to train the different components in CARE. It is important to note that the
role of CARE is to learn a representation that enables the incorporation of
metadata and functional abstraction. For end-to-end reinforcement
learning, it must be paired with a policy optimization algorithm. In the scope of this work, we use Soft Actor-Critic (SAC, \cite{soft-actor-critic}), but CARE can be paired with any policy optimization method.

\subsection{Incorporating Information from Metadata}

In the BC-MDP setting, the different tasks share objects and skills across the family of MDPs, but the state spaces across tasks are context-dependent, and therefore not the same, $\mathcal{S}^c$. Our goal is to reconstruct the universal state space $\mathcal{S}$. Of course, we do not have access to the true state space, but a useful inductive bias for the learning agent would be to learn composable representations for objects and attend over these representations for different tasks. One major challenge to this approach is that knowing which objects are relevant for each task requires object-level supervision, which is not commonly available. We propose to sidestep this problem by conditioning the attention on the task context, which is modelled using the easily available task metadata or description. Note that this metadata can be high-level, under-specified, and unstructured. It does not have to explain ``how to perform the task''; it can simply describe the task. Even in cases where this metadata is not readily available, it can be easily constructed. An example of a task description from Meta-World could be ``Reach a goal position''. Another example is MuJoCo tasks from Deepmind Control~\citep{dmc} and OpenAI Gym~\citep{openai_gym}, where humans identify the tasks by descriptive names like ``HalfCheetah Run'' and ``Maze Solver Ant'', as opposed to task ids $\{1,2,..., n\}$.

Given such a high-level task description, we focus on the case where the task context is captured using pre-trained language models. Specifically, we use the Roberta model~\citep{roberta_a_robustly_optimized_bert_pretraining_approach} to obtain a $768$-dimensional representation of the task description. This representation is projected to a lower-dimensional space using feedforward layers, and the resulting representation is denoted as the context $z_{context}$. The context is used to condition the different components of the policy, as described below.

\begin{figure*}[t]
    \centering
    \includegraphics[width=.9\textwidth]{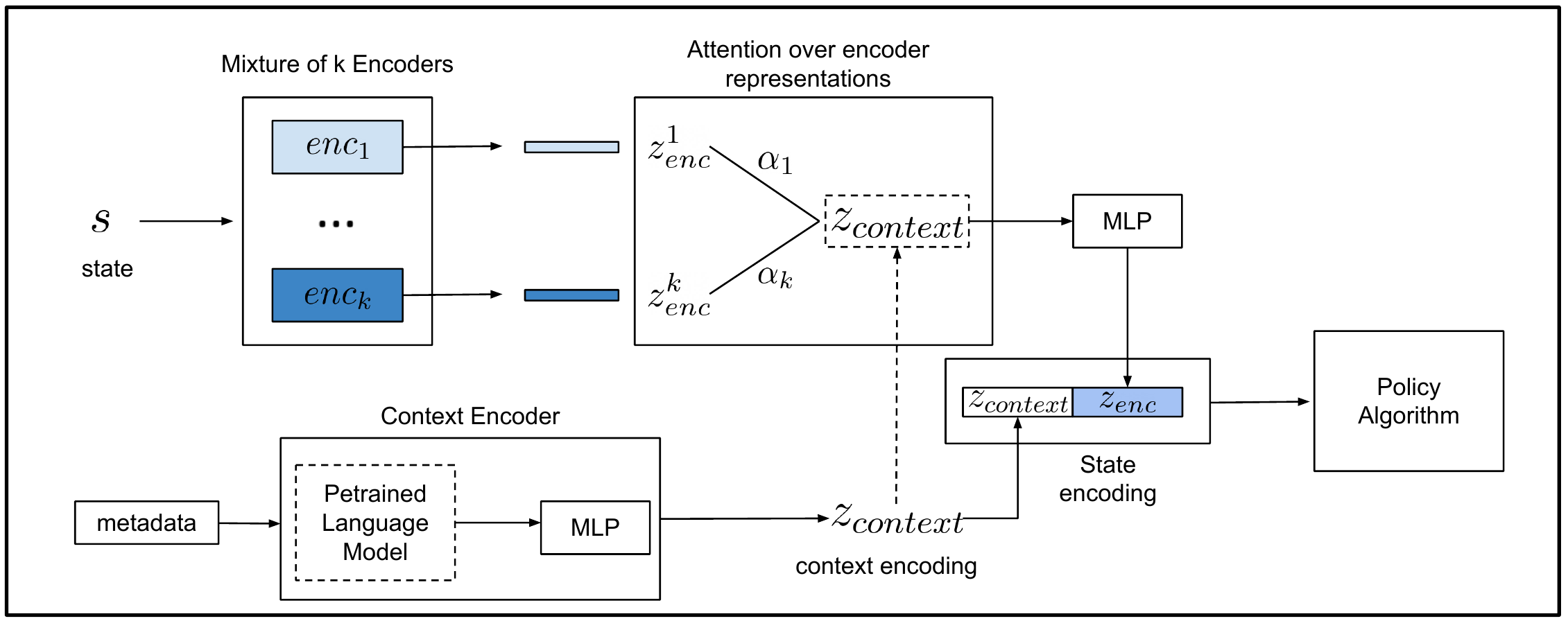}
    \caption{Architecture of the CARE model: Given an input state $s$, an mixture of $k$ encoders is used to compute $k$ encodings, denoted as $z_{enc}^{i} \forall i \in \{1, \cdots, k\}$. Given the metadata, a pretrained language model (followed by a feedforward network) is used to encode the task description as a real valued \textit{context} vector $z_{context}$. This context is used to compute an attention score between $z_{enc}^{i}$ and $z_{context}$. The attention scores are normalized (to sum to $1$) and are denoted as $\alpha_{i} \forall i \in \{1, \cdots, k\}$. The attention scores are used to compute a weighted sum of the encoders' representations and is denoted as $z_{enc}$. It is then concatenated with the context to obtain the state encoding $z_s$ that is given as input to the policy network. Dashed lines indicate that gradient does not flow through those components or computations. }
    \label{fig:architecture}
    \vspace{-10pt}
\end{figure*}

\setlength{\textfloatsep}{5pt}%
\begin{algorithm}[t]
\begin{algorithmic}[1]
\REQUIRE SAC Components 
\REQUIRE Context Encoder Network $C$
\REQUIRE $k$ Encoders $E_1, \dots, E_k$
\FOR{each timestep $n=1..N$}
    \FOR{each task $\task_i$}
        \STATE $z_{context}^i = C(metadata_i)$
        \STATE  $z_{enc}^j = E_j(s^i_{n}), \quad \forall j \in \{1, \dots, k\} $
        \STATE $\bar{z}_{context}^i = stopgrad(z_{context}^i)$
        \STATE  $\alpha_j = softmax(z_{enc}^j \boldsymbol{\cdot} \bar{z}_{context}^i) \quad \forall j \in \{1, ..., k\} $
        \STATE  $z_{enc}^{i} = MLP(\sum_{j \in \{1, \dots, k\}} z_{enc}^j \times \alpha_j$)
        \STATE  $z^{i} = z_{context}^{i}::z_{enc}^{i}$ (input to the SAC algorithm)
        \STATE \textsc{UpdateSAC}($\mathcal{D}, z^{i}$). Refer to~\cref{alg:sac}.
        \STATE \textsc{UpdateContextEncoder}($\mathcal{D}, z^{i}$). Refer to~\cref{alg:update_context_encoder}.
        \STATE \textsc{UpdateMixtureofEncoders}($\mathcal{D}, z^{i}$). Refer to~\cref{alg:update_moe}.
    \ENDFOR
\ENDFOR
\end{algorithmic}
\caption{{CARE algorithm for the multi-task RL setting.}}
\label{alg:care}
\end{algorithm}

\subsection{Contextual Attention based Representations}

We posit that a useful inductive bias for the training agent is to learn contextual representations for different tasks by learning multiple representations and attending over those representations using the task context. Given $N$ tasks, we use a mixture of $k$ encoders to learn $k$ state-representations. Here, $k$ is a hyperparameter and, in practice, it is much smaller than the number of tasks in the family of MDPs. Note that unlike some work on object-oriented learning, we do not assume access to \textit{privileged} information in terms of which objects are present in the input observation or useful to encode in a task. Moreover, while our design encourages the specialization of different encoders to combinations of different objects and skills, we note that this setup is incorporating a softer inductive bias than object-oriented learning~\cite{multi_object_representation_learning_with_iterative_variational_inference, object_centric_learning_with_slot_attention}. Specifically, while we are conditioning the policy on object representations, we do not explicitly model the interactions between the encoders/objects, as is done in recent works like~\citet{goyal2019recurrent}. We factorize the representation in terms of reusable components, unlike methods that first perform object detection and then model higher-order interactions between the objects using attention-based mechanisms or graph neural networks~\citep{kipf2018interacting,pathak2019LearningControlSelfAssembling,li2020causal}. We see our design choice as a trade-off between imposing more useful structure on the algorithm versus requiring less access to privileged information.

Given the $k$ encoders, we compute the $k$ representations $z_{enc}^{i} \forall i \in \{1, \cdots, k\}$. Given the context, $z_{context}$, we compute the normalized soft-attention weights for the encoder representations (denoted as $\alpha_i \forall i \in \{1, \cdots, k\}$). We pool the $k$ encoder representations into a fixed-size representation by performing a weighted sum using the \textbf{soft-attention} weights. The resulting encoder representation ($z_{enc}$) is computed as $z_{enc} = \sum_{i}^{k}\alpha_{i} \times z_{enc}^{i}$. We concatenate the encoder representation ($z_{enc}$) with the context representation ($z_{context}$) to obtain the state encoding ($z_s$). This state encoding is used as an input to the policy network, and the entire setup is trained end-to-end. Note that the language model is not updated during training and the context representation $z_{context}$ is detached from the computation graph before computing the attention weights. $z_{context}$ is updated using the policy loss directly, as it is a part of the state encoding. 

\subsection{Downstream Evaluation}
We use \textbf{Soft Actor-Critic} (SAC,(~\citet{soft-actor-critic} for downstream evaluation of the learned representations. SAC is an off-policy actor-critic method that uses the maximum entropy framework for soft policy iteration. At each iteration, SAC performs soft policy evaluation and improvement steps. For more details on SAC, refer \citet{soft-actor-critic}.

The overall algorithm is described in~\cref{alg:care}, with the sub-function details available in the Appendix (Algorithms~\ref{alg:sac},~\ref{alg:update_context_encoder},~\ref{alg:update_moe}). We note that steps 3 to 11 can be run concurrently for multiple tasks (as is done in our implementation). The architecture diagram is shown in \cref{fig:architecture} and the Appendix contains additional implementation details (\cref{app:implementation}) and hyper-parameters (\cref{app:hyperparameters}).

\section{Experiments}
\label{sec:experiments}

We now empirically evaluate the effectiveness of the proposed CARE model on Meta-World~\citep{meta-world} -- a multi-task RL benchmark with 50 tasks.
We design our experiments to answer the following questions: \textbf{i)} Is learning contextual attention-based representations an effective mechanism for knowledge transfer in multi-task RL? Does it perform better than methods that do not utilize this context? \textbf{ii)} Is the metadata useful only when learning compositional representations? \textbf{iii)} Does the metadata help to learn factored, specialized representations? \textbf{iv)} Does the metadata help in zero-shot generalization to unseen environments?

\subsection{How CARE compares to existing MTRL baselines}
\label{subsec::care_vs_baselines}

Existing works in multi-task RL come in two flavours: \textbf{i)} Extend a single task RL baseline for multi-task by using task-specific parameters (like one policy-head-per-task or per-task entropy regularization). These approaches are generally algorithm-agnostic. \textbf{ii)} Specialized multi-task algorithms. In this second category, we compare against the PCGrad~\citep{gradient_surgery_for_multitask_learning} algorithm which is specifically proposed for the Meta-World benchmark and achieves state-of-the-art results, outperforming multi-task algorithms like GradNorm~\citep{chen2018gradnorm} and Orthogonal Gradients (CosReg)~\citep{regularizing_deep_multi_task_networks_using_orthogonal_gradients}. We also compare with the Soft Modularization approach~\citep{multi_task_reinforcement_learning_with_soft_modularization} that performs routing in a shared policy network to learn different policies for different tasks and provides state-of-the-art results for Meta-World benchmark. Additionally, we compare with a popular and general-purpose conditioning method called FiLM~\cite{film}. We use FiLM layers to condition the encoder on the context (generated using a context encoder, just like in CARE). While FiLM is not a multi-task RL algorithm, it is used effectively in the language-conditioned RL setups~\cite{babyai, rtfm}.

The evaluation performance of the agent is computed as follows: At regular intervals, the agent is evaluated 5 times on each test environment, and the mean of the 5 runs is taken as the success rate for the corresponding environment. These success rates are averaged across the environments to obtain the mean success rate at every interval. A time-series of the mean success rates is obtained by evaluating the agent at regular intervals. The agent is trained for multiple seeds (10 in our case) resulting in 10 different time-series (one per seed) of mean success rates. These 10 time-series are averaged to compute the mean of the success rates (mean over the seeds). The best mean (across the time-series) is reported as the evaluation performance.

We highlight some challenges in comparing the performance of different models on the Meta-World suite. Generally, RL agents are evaluated for continuous-valued episodic rewards~\citep{openai_gym, dmc} while Meta-World uses a binary-valued success signal. While the use of success signals is not entirely unheard of~\cite{babyai}, we find that with Meta-World, we can improve the agent's performance just by increasing the frequency of evaluation (as shown in~\cref{table::mt10::care::1Keval} in Appendix). Evaluating the agent more often makes it more likely for the agent to solve a given task, making it harder to compare results across different works. For example, consider the extreme case where the agent is evaluated after every single update. Since we report the best (max) of the mean(success), evaluating more frequently could improve the performance since we are computing the max over a larger set. Thus, evaluation frequency acts like an implicit hyperparameter. We control for this issue in our setup by evaluating every agent at a fixed frequency (once every 10K environment steps, per task). Second, we find that the number of seeds (for evaluation) plays a big role in a model's reported performance. For example, we report that for MT10, the standard Multi-headed SAC achieves a mean success of 61\% (10 seeds), whereas~\citet{meta-world} reports a success rate of 88\% (1 seed), leading to a 44\% change in performance. Some other models~\citep{multi_task_reinforcement_learning_with_soft_modularization} use just 3 seeds. We account for the stochasticity of the evaluation process and ensure a fair comparison of all the models by running all the experiments with 10 seeds. We also report if our model's improvements are statistically significant or not. Additional details about testing for statistical significance can be found in~\cref{app::testing_for_statstical_significance}.

We use the same setup as other works. At regular interval, agent is evaluated on all envs

\begin{table}
\caption{Evaluation performance on the \textbf{MT10} test environments, after training for \textbf{2 million} steps (for each environment). Results are averaged over 10 seeds.  ME = Mixture of Encoders. CARE's improvement is statistically significant for baselines with *.}
\label{table::mt10::care}
\begin{tabular}{p{5.6cm}p{2.1cm}}
\toprule
         Agent & success \\
         & \small{(mean $\pm$ stderr)} \\
\midrule

    Multi-task SAC \cite{meta-world} * &        $0.49 \pm     0.073$ \\
    Multi-task SAC + Task Encoder * &      $0.54 \pm      0.047$ \\
    Multi-headed SAC \cite{meta-world} * &    $0.61 \pm      0.036$ \\
     PCGrad \cite{gradient_surgery_for_multitask_learning} * &         $0.72 \pm       0.022$ \\
     Soft Modularization \cite{multi_task_reinforcement_learning_with_soft_modularization} &        $0.73 \pm       0.043$ \\
     SAC + FiLM~\cite{film} &        $0.75 \pm       0.037$ \\
     SAC + Metadata + ME (\textbf{CARE}) &     $\mathbf{0.84} \pm       \mathbf{0.051}$ \\
     \midrule
    One SAC agent per task (upper bound) &        $0.90 \pm       0.032$ \\
\bottomrule
\end{tabular}
\vspace{5pt}
\end{table}

\begin{figure*}[t]
    \centering
    \begin{subfigure}[b]{0.30\textwidth}
        \includegraphics[width=\textwidth]{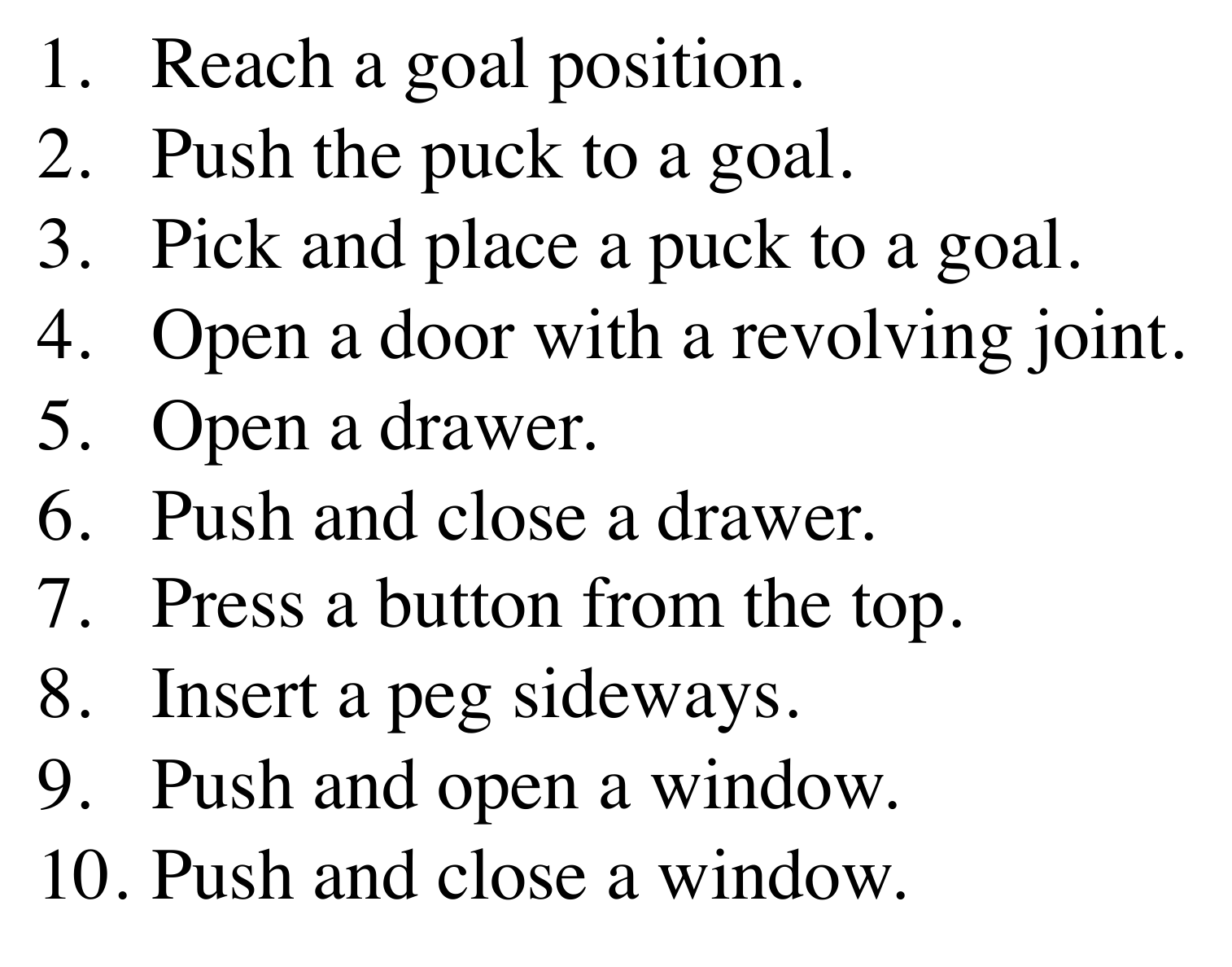}\hfill
        \caption{}
        \label{fig:similarity_viz:task_description}
    \end{subfigure}\hspace{-5pt}
    \begin{subfigure}[b]{0.22\textwidth}
        \includegraphics[width=\textwidth]{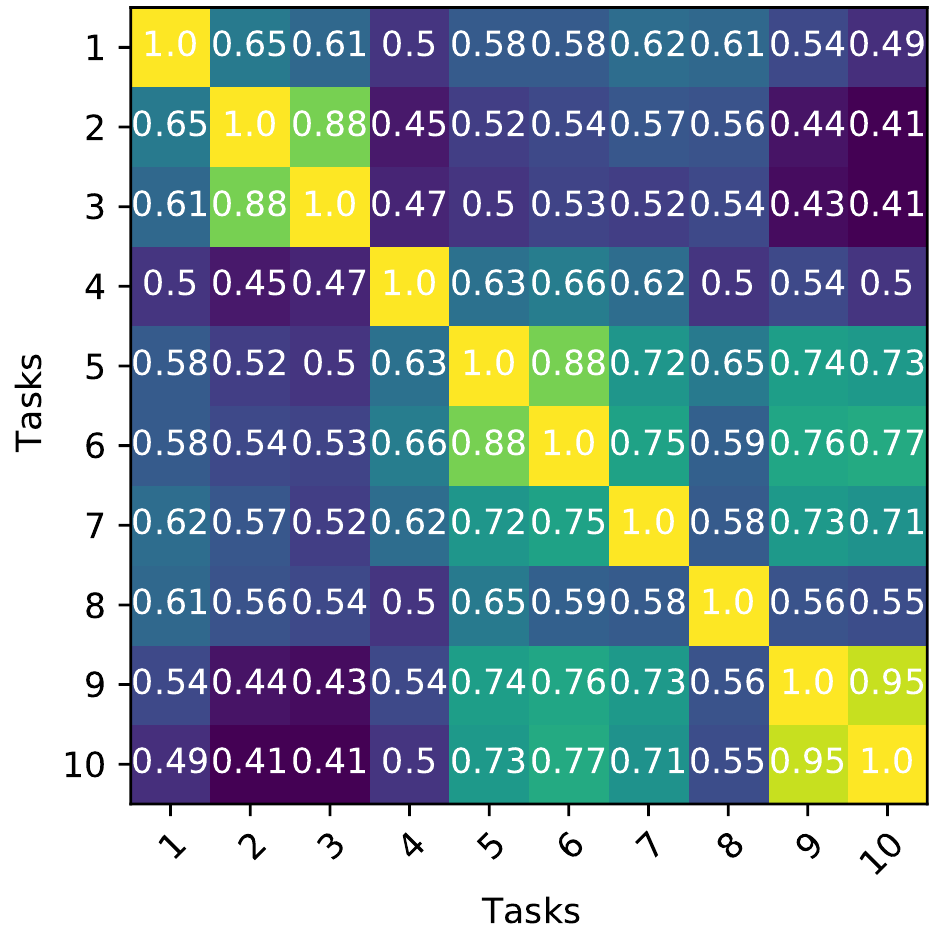}\hfill
        \caption{}
        \label{fig:similarity_viz:pretrained}
    \end{subfigure}    \hspace{-5pt}
    \begin{subfigure}[b]{0.22\textwidth}
        \includegraphics[width=\textwidth]{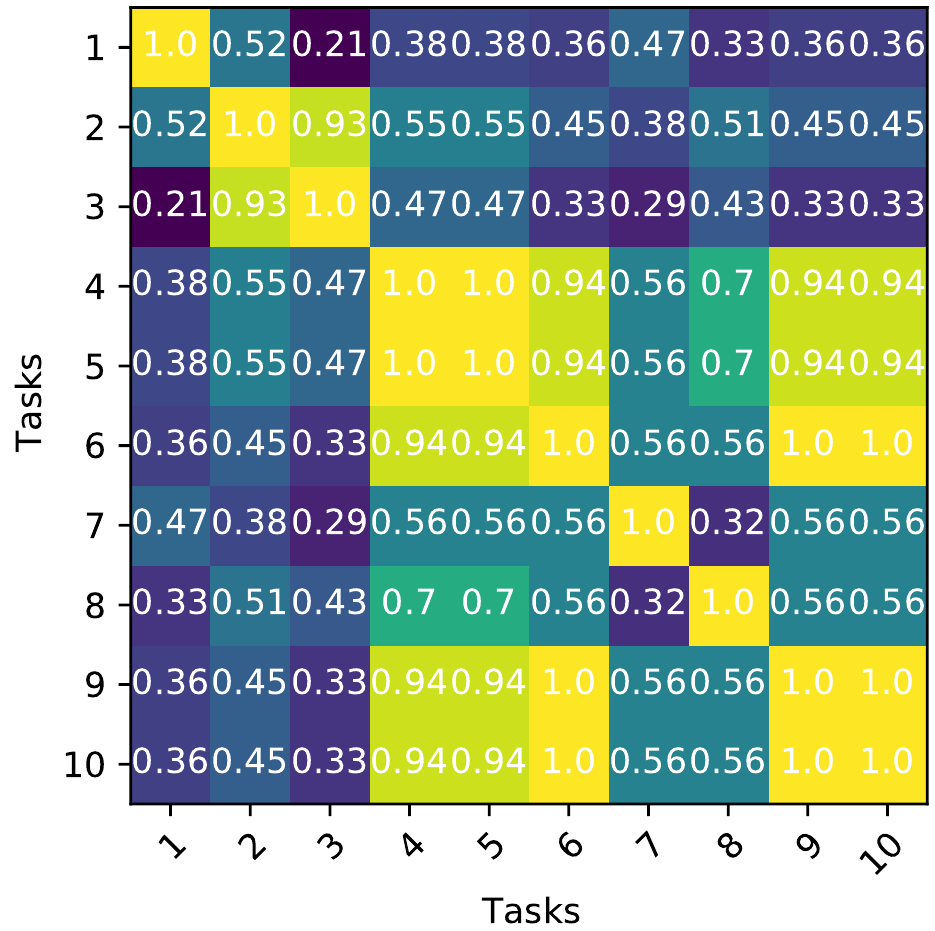}\hfill
        \caption{}
        \label{fig:similarity_viz:care}
        \end{subfigure}    \hspace{-5pt}
    \begin{subfigure}[b]{0.22\textwidth}
    \includegraphics[width=\textwidth]{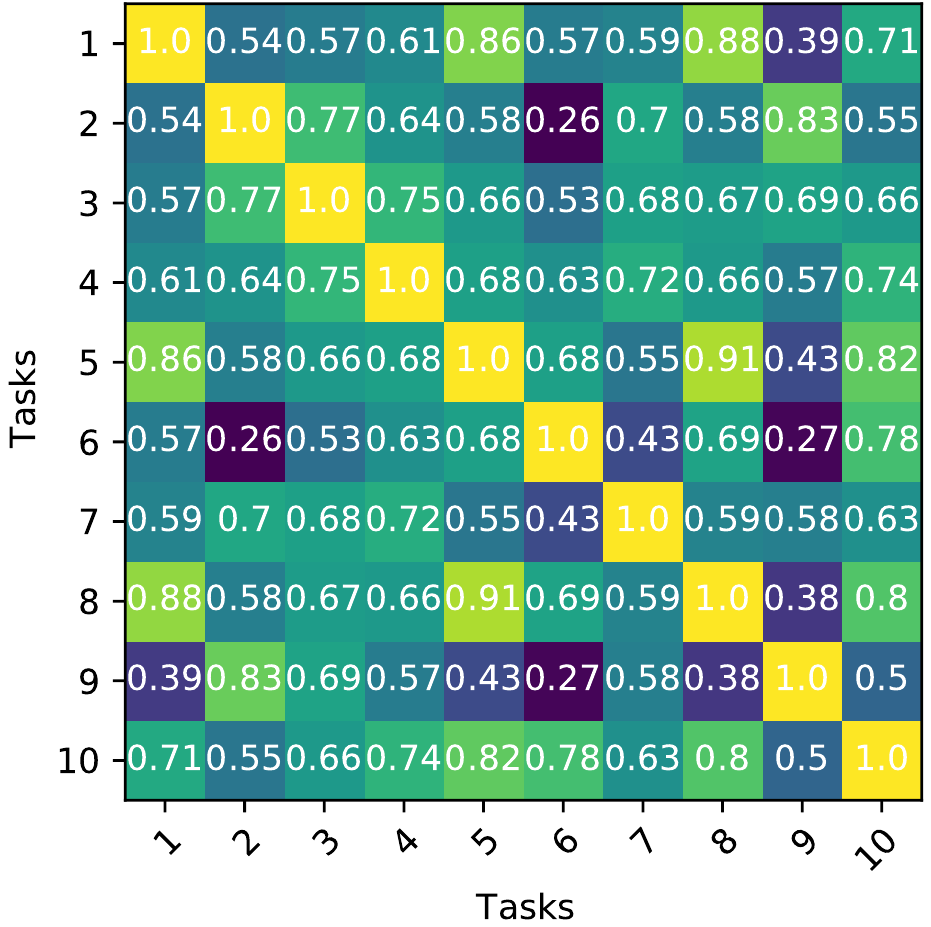}\hfill
    \caption{}
        \label{fig:similarity_viz:no_metadata}
    \end{subfigure}    
    \caption{(a): Task descriptions for MT10. (b) Cosine similarity (for MT10) between the pretrained task embeddings. (c) Cosine similarity (for MT10) between the context representations using CARE model with $k=6$ encoders. (d) Cosine similarity (for MT10) between the context representations without using metadata with $k=6$ encoders. Structure across tasks is clearly exhibited in (b) and (c), but not in (d), which has no access to metadata.}
    \label{fig:similarity_viz}
    \vspace{-15pt}
\end{figure*}
\begin{table}[t]
\centering
\vspace{-10pt}
\caption{Evaluation performance on the \textbf{MT10} test environments, after training for \textbf{100 thousand} steps (for each environment). Results averaged over 10 seeds.  ME = Mixture of Encoders. CARE's improvement is statistically significant for baselines with *.}
\label{table::mt10::care_100k}
\begin{tabular}{p{5.6cm}p{2.1cm}}
\toprule
         Agent & success \\
         & \small{(mean $\pm$ stderr)} \\
\midrule

    Multi-task SAC \cite{meta-world} * &        0.13 $\pm$     0.022 \\
    Multi-task SAC + Task Encoder * &      0.14 $\pm$  0.012 \\
    Multi-headed SAC \cite{meta-world} * &    $0.17 \pm  0.033$ \\
     PCGrad \cite{gradient_surgery_for_multitask_learning}  * &        $0.20  \pm     0.032$ \\
     Soft Modularization \cite{multi_task_reinforcement_learning_with_soft_modularization} &         $0.33 \pm      0.036$ \\
     SAC + FiLM~\cite{film} * &        $0.27 \pm       0.037$ \\
     SAC + Metadata + ME ( \textbf{CARE}) &    $ \mathbf{0.36} \pm      \mathbf{0.035}$ \\
\bottomrule
\end{tabular}
\vspace{5pt}
\end{table}

\begin{table}[t]
\centering
\caption{Evaluation performance on the \textbf{MT50} test environments, after training for \textbf{2 million} steps (for each environment). Results are averaged over 10 seeds.  ME = Mixture of Encoders. CARE's improvement is statistically significant for baselines with *.}
\label{table::mt50::care}
\begin{tabular}{p{5.6cm}p{2.1cm}}
\toprule
         Agent & success \\
         & \small{(mean $\pm$ stderr)} \\
\midrule

    Multi-task SAC \cite{meta-world} * &        $0.36 \pm 0.013$ \\
    Multi-task SAC + Task Encoder * &      $0.40 \pm     0.024$ \\
    Multi-headed SAC \cite{meta-world}  &    $0.45 \pm      0.064$ \\
    PCGrad \cite{gradient_surgery_for_multitask_learning} &         $0.5 \pm  0.017$ \\
    Soft Modularization \cite{multi_task_reinforcement_learning_with_soft_modularization} &         $0.5 \pm    0.035$ \\
    SAC + FiLM \cite{film} * &  0.40 $\pm$  0.012 \\
    SAC + Metadata + ME (\textbf{CARE}) &     $\mathbf{0.54} \pm       \mathbf{0.031}$ \\
    \midrule
    One SAC agent per task (upper bound) &        $0.74 \pm       0.041$ \\
\bottomrule
\end{tabular}
\end{table}

\begin{table}[t]
\centering
\caption{Evaluation performance on the \textbf{MT50} test environments, after training for \textbf{100 thousand} steps (for each environment). Results averaged over 10 seeds.  ME = Mixture of Encoders. CARE's improvement is statistically significant for baselines with *.}
\label{table::mt50::care_100k}
\begin{tabular}{p{5.9cm}p{2.1cm}}
\toprule
         Agent & success \\
         & \small{(mean $\pm$ stderr)} \\
\midrule

    Multi-task SAC \cite{meta-world}  * &        $0.13 \pm   0.0061$ \\
    Multi-task SAC + Task Encoder * &      $0.28 \pm   0.015$ \\
    Multi-headed SAC \cite{meta-world} * &    $0.19 \pm 0.0071$ \\
     PCGrad \cite{gradient_surgery_for_multitask_learning} * &        $0.21  \pm    0.0068$ \\
     Soft Modularization~\cite{multi_task_reinforcement_learning_with_soft_modularization} * &         $0.20 \pm   0.023$ \\
     SAC + FiLM \cite{film} * &  0.16 $\pm$  0.006 \\
     SAC + Metadata + ME ( \textbf{CARE}) &     $\mathbf{0.40} \pm   \mathbf{0.015}$ \\
\bottomrule
\end{tabular}
\end{table}

We note that these challenges are not inherent limitations of Meta-World and issues related to seeds affect the evaluation of RL algorithms in general~\citep{deep_reinforcement_learning_that_matters}. However, we believe that some of these challenges can be alleviated by standardizing the evaluation protocol. Given the usefulness of Meta-World as a multi-task RL benchmark, these challenges should be highlighted to ensure they are considered in the subsequent works.

Meta-World benchmark provides two setups - \textbf{MT10}: a suite of 10 tasks and \textbf{MT50}: a suite of 50 tasks (a superset of MT10). We use both setups for evaluation.
In~\cref{table::mt10::care}, we compare the performance of the CARE model for MT10 with the different baselines and report the performance after 2M steps. Following the setup in~\citet{kaiser2019model,srinivas2020curl}, we also compare the performance in the low-sample regime with 100K steps per task in~\cref{table::mt10::care_100k} and 500K steps per task in~\cref{table::mt10::care_500k} (in the Appendix). We note that the proposed CARE model consistently outperforms the other models for the MT10 task. We also note that among specialized multi-task algorithms, PCGrad~\citep{gradient_surgery_for_multitask_learning} performs quite poorly in the low-sample regime.

Similarly, in~\cref{table::mt50::care} and~\cref{table::mt50::care_100k}, we compare the performance of the models for MT50 after 2M and 100K steps respectively. The corresponding table for 500K steps is in the Appendix (Table~\ref{table::mt50::care_500k}). We note that not only does CARE outperform the other baselines, it is much more sample efficient than other baselines in the low-sample regime.

\begin{table}[ht]
\centering
\vspace{-10pt}
\caption{Effect of using the metadata or the ensemble of encoders or both (proposed CARE model). Evaluation performance on the MT10 test environments, after training for \textbf{2 million} steps (for each environment). ME = Mixture of Encoders. CARE's improvement is statistically significant for the baselines marked with *.}
\label{table::mt10::ablation_with_metadata_and_ensemble}
\resizebox{0.5\textwidth}{!}{%
\begin{tabular}{p{5.4cm}p{2.1cm}}
\toprule
         Agent & success \\
         & \small{(mean $\pm$ stderr)} \\
\midrule
    SAC + ME &      $0.74 \pm 0.043$ \\
    SAC + Metadata &      $0.79 \pm     0.041$ \\
    SAC + Metadata + ME (\textbf{CARE}) &     $\textbf{0.84} \pm       \mathbf{0.051}$ \\
\bottomrule
\end{tabular}
}
\end{table}

One additional benefit of the CARE model is that it can be easily combined with more powerful policies and learning algorithms. For example, we find that using multi-headed SAC with the CARE model significantly improves the performance on the MT50 setup (mean success of $0.61$, with a standard error of $0.0287$). Since the CARE model focuses on learning representations, it can benefit from the improvements in policy optimisation for multi-task RL.

In~\cref{app::results::mt10_ablations_with_encoders} we consider some ablations with the CARE model. First, we vary the number of encoders, showing that increasing the number of encoders hurts performance when shared information is no longer leveraged. We also try using only  top-$k$ encoders (i.e. encoders with top-$k$ highest attention scores) with hard attention, confirming the robustness of our method to different aggregation techniques. Finally, we design an experiment where we hardcode the mapping between the tasks and the encoders, showing that mapping encoders to specific objects and skills helps performance.

\begin{table}[ht]
\centering
\vspace{-5pt}
\caption{Effect of using the metadata or the ensemble of encoders or both (proposed CARE model). Evaluation performance on the MT50 test environments, after training for \textbf{2 million} steps (for each environment). ME = Mixture of Encoders. CARE's improvement is statistically significant for the baselines marked with *.}
\label{table::mt50::ablation_with_metadata_and_ensemble}
\resizebox{0.5\textwidth}{!}{%
\begin{tabular}{p{5.4cm}p{2.1cm}}
\toprule
         Agent & success \\
         & \small{(mean $\pm$ stderr)} \\
\midrule
    SAC + Mixture of Encoders * &      $0.44 \pm  0.012$ \\
    SAC + Metadata &      $0.48 \pm  0.025$ \\
    SAC + Metadata + ME (\textbf{CARE}) &     $\mathbf{0.54} \pm       \mathbf{0.031}$ \\
\bottomrule
\end{tabular}}
\end{table}

\begin{table}
\centering
\caption{Evaluation performance on held-out environments from MT10, after training on remaining 8 environments for \textbf{2 million} steps for each environment. Results averaged over 10 seeds. ME = Mixture of Encoders. CARE's improvement is statistically significant for baselines marked with *.}
\label{table::mt10::heldout}
\begin{tabular}{p{5.6cm}p{2.1cm}}
\toprule
         Agent & success \\
         & \small{(mean $\pm$ stderr)} \\
\midrule

    PCGrad \cite{gradient_surgery_for_multitask_learning} * & 0.05 $\pm$ 0.076 \\
    Soft Modularization \cite{multi_task_reinforcement_learning_with_soft_modularization} & 0.1 $\pm$  0.089 \\
    SAC + FiLM \cite{film} &     0.2 $\pm$ 0.073        \\
    SAC + Metadata + ME (\textbf{CARE}) &  \textbf{0.3} $\pm$ \textbf{0.077} \\
\bottomrule
\end{tabular}
\end{table}
\subsection{Is the metadata useful only when learning compositional representations?}
In \cref{subsec::care_vs_baselines}, we showed that the metadata is useful for learning compositional representations in the CARE model. A follow-up question is whether the metadata is also useful when using a single encoder. We address this question in Tables~\ref{table::mt10::ablation_with_metadata_and_ensemble} and~\ref{table::mt50::ablation_with_metadata_and_ensemble} where we compare the performance of the CARE model in the absence of metadata or in the absence of mixture of encoders. We note that removing the metadata (first row) hurts the models more than removing the mixture of encoders (second row). While using metadata with just a single encoder (second row) provides comparable performance to the other baselines, using it with a mixture of encoders leads to even better performance.

\vspace{-5pt}
\subsection{Interpreting the specialized representations}

We perform a visual investigation of what information is being shared across tasks and the role of metadata in the information sharing. Specifically, in \cref{fig:similarity_viz} we show cosine similarity between the context representations under the 10 tasks of MT10. For each task, we feed the context through the pretrained language model (RoBERTa) and refer to the resulting representation as the pretrained task embedding. In~\cref{fig:similarity_viz:pretrained}, we show the cosine similarity between just these pretrained task embeddings, which are the input to the MLP of the context encoder (\cref{fig:architecture}). In~\cref{fig:similarity_viz:care}, we show the cosine similarity between the context representations from the proposed CARE model (with 6 encoders) after training. We observe a similar structure in both similarity matrices, that tasks with semantically similar descriptions also have similar context encodings. For example, task 2 and 3 require the agent to interact with the same object. Similarly tasks 4, 5, and 6 are related by the skill ``open'' and object ``drawer''. We now want to observe if these similarities can be found from just the tasks, without access to metadata. In~\cref{fig:similarity_viz:no_metadata}, we plot the cosine similarity between the context representations from the CARE model (with 6 encoders) without the use of task metadata. Some similarities are present, but nothing as strongly correlated as seen in \cref{fig:similarity_viz:care}. This result further supports our hypothesis that task metadata plays an important role in inferring similarity between tasks, and shows the interpretability of the CARE representation.

\vspace{-5pt}
\subsection{Zero-shot generalization to unseen environments}
Given the compositionality present in language, we want to evaluate if CARE can be used for zero-shot generalization to unseen environments. We train the agents on 8 environments from MT10 and evaluate on two held-out environments, ``drawer-open-v1'' and ``window-open-v1''. There are three training environments that directly relate to these test environments, ``drawer-close-v1'', ``window-close-v1'', ``door-open-v1'', thus allowing for the possibility of zero-shot generalization. In~\cref{table::mt10::heldout}, we observe that CARE exhibits some promising performance. We note that the comparison is unfair to PCGrad and Soft Modularization as they do not have any means for generalizing to the unseen task, but our motivation is to highlight the potential benefits of using metadata for zero-shot generalization to unseen environments. We further note that CARE generalizes better than FiLM, which also leverages metadata.

\section{Related Work}

Multi-task learning holds the promise of accelerating learning across multiple tasks by sharing useful information~\citep{caruana1997multitask_learning, zhang2014facial_landmark_detection_by_deep_multitask_learning, kokkinos2017ubernet, radford2019language_models_are_unsupervised_multitask_learners, epopt_learning_robust_neural_network_policies_using_model_ensembles, ruder2017overview, end_to_end_multi_task_learning_with_attention, mott2019towards, electronics9091363}. \textbf{Multi-task reinforcement learning} (MTRL) has been extensively studied with the focus on assumptions around shared properties and structures of different tasks.
\cite{calandriello2014multitask,borsa2016mtrl,maurerBenefitMultitaskRepresentation} assume that in the multi-task settings, tasks share a common, low dimensional representation, and therefore advocate for learning a shared representation space across all the tasks.~\cite{bram2019attentive} considered the setup when the action space between different tasks is not aligned.  \citet{zhang2020hipbmdp} describes multi-task learning algorithms where the tasks have different dynamics but a shared reward structure and makes the assumption of a universal dynamics model. All of these methods for the MTRL setting only utilize a simplistic context in the form of an ordinal task id. Richer contexts in the form of task embeddings are learned online from the differences in reward and dynamics across tasks. We instead focus on the setting where side information is available and can be utilized as a richer context than task ids.

Several works have focused on the problem of \textbf{negative interference}~\citep{adapting_auxiliary_losses_using_gradient_similarity, regularizing_deep_multi_task_networks_using_orthogonal_gradients, gradient_surgery_for_multitask_learning} where the gradients corresponding to the different tasks interfere negatively with each other. Along with slowing down training, conflicting gradients could cause the agent to \textit{unlearn} knowledge of one task to learn another task. Despite some successes, the proposed approaches are unsatisfactory, either because they increase the computational/memory overhead (for example, \citet{gradient_surgery_for_multitask_learning} introduces an $\mathcal{O}(n^2)$ complexity, where $n$ is the number of tasks) or because they require the training model to ignore some components of the gradients, thus slowing down learning and deteriorating sample efficiency. We propose to use the context for deciding which information should be shared across tasks, thus alleviating negative interference.

\textbf{Contextual MDPs} have been previously defined and analyzed as a setting where side information is exploited for transfer across tasks~\citep{hallak2015contextual,modi_markov_2017}. \citet{hallak2015contextual} first defined the contextual MDP setting, drawing connections to contextual multi-arm bandits~\citep{lai1985mab, langford2007mab}. However, they assume that the state spaces across contexts are the same.
\citet{modi_markov_2017} requires an assumption of smoothness in the MDP parameters with respect to the context and examine the online learning scenario, providing an extension of the Rmax algorithm with PAC bounds. They also assume that the context is given. 
\citet{klink2019contextualrl} adopt the contextual MDP setting but with an additional assumption that the agent can control the context, and therefore the task distribution. They propose an algorithm to generate a curriculum that allows the agent to gradually progress to a target context distribution. 
In our work, we relax several of the assumptions made by these prior works and extend the type of context explored.

\textbf{Multi-task learning with metadata} has been used in \citet{Mmtadata_based_clustered_multitask_learning_for_thread_mining_in_web_communities, zheng2019metadata} where the focus is on Task Relation Discovery in the context of supervised learning. In Reinforcement Learning, our work has close ties to \textbf{language-conditioned RL}, where natural language phrases have been used as part of task descriptions in the context of several single task RL setups like goal-oriented RL~\citep{chao2011towards, babyai}, grounded language acquisition~\citep{grounded_language_learning_in_a_simulated_3d_world, gated_attention_architectures_for_task_oriented_language_grounding}, and instruction following~\citep{understanding_natural_language_commands_for_robotic_navigation_and_mobile_manipulation, learning_to_interpret_natural_language_navigation_instructions_from_observations, learning_to_parse_natural_language_to_grounded_reward_functions_with_weak_supervision}. Similar to our BC-MDP setting, recent work~\citep{rtfm} also defines meta-environments specific to the language-conditioned setting. 
In all these works, the language description is a part of the problem specification and not just extra information, while our work proposes a way to incorporate auxiliary side information to improve sample efficiency and generalization in multi-task RL. Many of these works make additional assumptions about the language description. For example,~\cite{shu2017hierarchical} uses ``a two-word tuple template consisting of a skill and an item'' to describe the task while the metadata for CARE can be high-level, under-specified, and unstructured. Further, we note that the contextual setting we propose includes information beyond just text, and can also include images or features.

Several works have also focused on \textbf{learning compositional models} for multi-task learning. \citet{end_to_end_multi_task_learning_with_attention} trains a network with task-specific soft-attention modules.~\citet{learning_modular_neural_network_policies_for_multi_task_and_multi_robot_transfer} decompose the policy into two components -- ``task-specific'' (shared across all robots) and ``robot-specific'' (shared across all tasks). \citet{automatically_composing_representation_transformations_as_a_means_for_generalization} consider a family of algorithmic tasks, with different levels of complexity. \citet{multi_task_reinforcement_learning_with_soft_modularization} performs routing in a base policy network to generate different policies for different tasks. Unlike our work, these works do not leverage metadata to learn a context on which the module decomposition can be conditioned on.

\section{Discussion}
In this work, we highlight an under-explored setting of using contextual information to improve performance in multi-task reinforcement learning.  We show that metadata, which is often present in multi-task settings, can be leveraged within the MDP framework to improve performance by enabling more efficient sharing of information across tasks. Further, we define a new setting, the block contextual Markov decision process (BC-MDP), to handle settings where the state space can differ across tasks. Finally, we show that a mixture of encoders can effectively learn a context-dependent representation that can be used with a single policy to solve a family of tasks. We showcase our method, CARE, on Meta-World and achieve new state-of-the-art results.

There are two dimensions along which this work can be extended. Here, we only explore a specific type of context in the form of text descriptions for each task. Other forms of context can also be considered, such as people or places. As examples, in personalized medicine or recommendation systems, the user can be used as a form of context to adapt and share knowledge from other users to transfer the policy. Similarly, in household robotics, the location and layouts of homes will differ although the task is the same. 

The relevance of this work also carries over to rich observation settings, or the high-dimensional version of the block contextual MDP setting we examine here~\citep{mozifian2020intervention}. This setting is relevant in robotics, where the agent typically has a first person view of the world. The entire state space (the world) cannot be captured in each task. Instead, the agent only focuses on a relevant subspace of the entire state space at a time. This extension significantly enriches the set of multi-task problems we can tackle and brings us closer to the intractable partial observability (POMDP) setting~\citep{kaelbling1998pomdp,zhang2019causal_states}, but offers more flexibility that allows for additional tractability by limiting the partial observability settings we consider.

\section*{Acknowledgements}

We thank Edward Grefenstette, Tim Rockt\"{a}schel, Danielle Rothermel and Olivier Delalleau for feedback that improved this paper.

\pagebreak

\bibliography{icml}
\bibliographystyle{icml2021}

\clearpage

\appendix
\newcommand{\policy}{\pi}
\newcommand{\policyold}{{\policy_\mathrm{old}}}
\newcommand{\policynew}{{\policy_\mathrm{new}}}
\newcommand{\policyme}{\pi_\mathrm{me}}
\newcommand{\policyopt}{{\policy^*}}
\newcommand{\paramsz}{{\theta_0}}

\newcommand{\paramsi}{{\theta_i}}
\newcommand{\paramsip}{{\theta_{i+1}}}
\newcommand{\softpolicy}{{\policy_\mathrm{soft}}}
\newcommand{\hardpolicy}{{\policy^{\dagger}}}

\newcommand{\qparams}{{\theta}}   %
\newcommand{\eparams}{{\phi}}   %

\newcommand{\qtargetparams}{{\bar\theta}}   %

\newcommand{\latspace}{\mathcal{H}}
\newcommand{\sz}{{\state_0}}
\newcommand{\stm}{{\state_{t-1}}}
\newcommand{\st}{{\state_t}}
\newcommand{\sti}{{\state_t^{(i)}}}
\newcommand{\sT}{{\state_T}}
\newcommand{\stp}{{\state_{t+1}}}
\newcommand{\stpi}{{\state_{t+1}^{(i)}}}
\newcommand{\density}{p}
\newcommand{\pdyn}{\density}

\newcommand{\action}{\mathbf{a}}
\newcommand{\at}{{\action_t}}
\newcommand{\V}{V}
\newcommand{\Q}{Q}
\newcommand{\Expectation}[2]{\operatorname{\mathbb{E}}_{#1}\left[#2\right]}
\newcommand{\pparams}{{\phi}}   %
\newcommand{\params}{\theta}
\newcommand{\vparams}{{\psi}}   %
\newcommand{\vtargetparams}{{\bar\psi}}   %
\newcommand{\reward}{r}
\newcommand{\discount}{\gamma}
\newcommand{\kl}[2]{\mathrm{D_{KL}}\left(#1\;\middle\|\;#2\right)}
\newcommand{\voidarg}{{\,\cdot\,}}

\section{Additional Implementation Details}
\label{app:implementation}

\subsection{Libraries}

We use the following open-source libraries: PyTorch~\cite{2019pytorch}\footnote{\url{https://pytorch.org/}}, Hydra~\cite{Yadan2019Hydra}\footnote{\url{https://github.com/facebookresearch/hydra}}, MetaWorld~\cite{meta-world}\footnote{\url{https://github.com/rlworkgroup/metaworld}}, MTEnv~\cite{Sodhani2021MTEnv}\footnote{\url{https://github.com/facebookresearch/mtenv}}, MTRL~\cite{Sodhani2021MTRL}\footnote{\url{https://github.com/facebookresearch/mtrl}}, Numpy~\cite{harris2020array}\footnote{\url{https://numpy.org/}} and Pandas~\cite{reback2020pandas}\footnote{\url{https://pandas.pydata.org/}}. For MetaWorld, we use the following commit-id to run our experiments: \url{https://github.com/rlworkgroup/metaworld/commit/af8417bfc82a3e249b4b02156518d775f29eb289}

\subsection{SAC Algorithm}

We use the SAC policy algorithm~\cite{soft-actor-critic} to learn representation for the CARE model. We provide the pseduo-code for SAC, along with the key equations. For more details, refer to~\citet{soft-actor-critic}.

\begin{algorithm}[ht]
\begin{algorithmic}[1]
\REQUIRE SAC components i.e. Policy parameters ($\pparams$), Value function parameters ($\vparams$), Target value function parameters ($\vtargetparams$) and Q-function parameters ($\params$)
\STATE  $a_t \sim \pi(\cdot | z^i_{t})$
\STATE Execute $a_t$ in the environment (for task $\task_i$) and get the next state $s_{t+1}$, reward $r_{t}$ and \textit{done} signal $d_{t}$.
\STATE $\mathcal{D} \leftarrow \mathcal{D} \cup (s_t, a_t, s_{t+1}, r_{t}, d_{t}, i)$
\STATE $\vparams \leftarrow \vparams - \lambda_V \hat \nabla_\vparams J_\V(\vparams)$ (\cref{eq:v_gradient})
\STATE $\vtargetparams\leftarrow \tau \vparams + (1-\tau)\vtargetparams$
\STATE $\params_i \leftarrow \params_i - \lambda_Q \hat \nabla_{\params_i} J_\Q(\params_i)$ for $i\in\{1, 2\}$ (\cref{eq:q_gradient})
\STATE $\pparams \leftarrow \pparams - \lambda_\policy \hat \nabla_\pparams J_\policy(\pparams)$ (\cref{eq:policy_objective})
\end{algorithmic}
\caption{\textsc{UpdateSAC}($\mathcal{D}, z^{i}_{t}$)}
\label{alg:sac}
\end{algorithm}

The soft value function is trained to minimize the squared residual error 
\begin{align}
\label{eq:v_cost}
\resizebox{\columnwidth}{!}{$
J_V(\vparams) = \Expectation{\st \sim \mathcal{D}}{\frac{1}{2}\left(\V_\vparams(\st) - \Expectation{\at\sim\policy_\pparams}{Q_\params(\st, \at) - \log \policy_\pparams(\at|\st)}\right)^2}\,$}.
\end{align}

The gradient of \cref{eq:v_cost} is estimated with an unbiased estimator
\begin{align}
\resizebox{\columnwidth}{!}{$
\hat \nabla_\vparams J_V(\vparams) = \nabla_\vparams \V_\vparams(\st) \left(\V_\vparams(\st) - Q_\params(\st, \at) + \log \policy_\pparams(\at|\st)\right),$}
\label{eq:v_gradient}
\end{align}
where the actions are sampled according to the current policy, instead of the replay buffer.  The update uses a target value network $V_{\bar\psi}$, where $\bar\psi$ is an exponentially moving average of the value network weights, which has been shown to stabilize training~\citep{mnih2015human}.

The soft Q-function parameters are trained to minimize the soft Bellman residual
\begin{align}
J_\Q(\params) = \Expectation{(\st, \at)\sim\mathcal{D}}{\frac{1}{2}\left(\Q_\params(\st, \at) - \hat \Q(\st, \at)\right)^2},
\label{eq:q_cost}
\end{align}
with 
\begin{align}
\hat \Q(\st, \at) = \reward(\st, \at) + \discount \Expectation{\stp\sim\pdyn}{\V_\vtargetparams(\stp)},
\end{align}
which is optimized with stochastic gradients
\begin{align}
\resizebox{\columnwidth}{!}{$
\hat \nabla_\params J_Q(\params) =  \nabla_\params \Q_\params(\at, \st) \left(\Q_\params(\st, \at) - \reward(\st, \at) - \discount \V_\vtargetparams(\stp)\right)$}.
\label{eq:q_gradient}
\end{align}

The policy parameters are learned by minimizing:
\begin{align}
J_\policy(\pparams) = \Expectation{\st\sim\mathcal{D}}{\kl{\policy_\pparams(\voidarg|\st)}{\frac{\exp\left(Q_\params(\st, \voidarg)\right)}{Z_\params(\st)}}}.
\label{eq:policy_objective}
\end{align}

\subsection{CARE Components}

The CARE algorithm uses two components: (i) A context encoder network and (ii) mixture of $k$ encoders (shown in~\cref{fig:architecture}). Both components are trained using the policy loss as described in~\cref{alg:update_context_encoder} and~\cref{alg:update_moe} respectively. We note that for computing the encoder representation $z_{enc}$, the context encoding $z_{context}$ is detached from the computational graph.

\begin{algorithm}[ht]
\begin{algorithmic}[1]
\REQUIRE Context Encoder Network $C$ with $\omega$ as the parameters, SAC components
\STATE  $a_t \sim \pi(\cdot | z^i_{t})$
\STATE Execute $a_t$ in the environment (for task $\task_i$) and get the next state $s_{t+1}$, reward $r_{t}$ and \textit{done} signal $d_{t}$.
\STATE $\mathcal{D} \leftarrow \mathcal{D} \cup (s_t, a_t, s_{t+1}, r_{t}, d_{t}, i)$
\STATE $J_{C}(\omega) = J_{V} + J_{Q} + J_{\pi}$ 
\STATE $\omega \leftarrow \omega - \lambda_\omega \hat \nabla_\omega J_{C}(\omega)$
\end{algorithmic}
\caption{\textsc{UpdateContextEncoder}($\mathcal{D}, z^{i}_{t}$)}
\label{alg:update_context_encoder}
\end{algorithm}

\begin{algorithm}[ht]
\begin{algorithmic}[1]
\REQUIRE $k$ Encoders $E_1, \dots, E_k$ with $\zeta_{k}$ as the parameters, SAC components
\STATE  $a_t \sim \pi(\cdot | z^i_{t})$
\STATE Execute $a_t$ in the environment (for task $\task_i$) and get the next state $s_{t+1}$, reward $r_{t}$ and \textit{done} signal $d_{t}$.
\FOR{each encoder in $k=1..K$} 
    \STATE $\mathcal{D} \leftarrow \mathcal{D} \cup (s_t, a_t, s_{t+1}, r_{t}, d_{t}, i)$
    \STATE $J_{E_{k}}(\zeta_{k}) = J_{V} + J_{Q} + J_{\pi}$ 
	\STATE $\zeta_{k} \leftarrow \zeta_{k} - \lambda_{\zeta_{k}} \hat \nabla_{\zeta_{k}} J_{E_k}(\zeta_{k})$
\ENDFOR

\end{algorithmic}
\caption{\textsc{UpdateMixtureOfEncoders}($\mathcal{D}, z^{i}_{t}$)}
\label{alg:update_moe}
\end{algorithm}

\section{Hyperparameter Details}
\label{app:hyperparameters}

In this section, we provide hyper-parameter values for each of the methods in our experimental evaluation. In~\cref{table::common_hp}, we provide the hyperparameter values that are common acrss all the methods.

\begin{table}[ht]
\centering
\caption{Hyperparameter values that are common across all the methods.}
\label{table::common_hp}
\resizebox{0.5\textwidth}{!}
{
    \begin{tabular}{p{3.7cm}p{3.7cm}}
    \toprule
         Hyperparameter & Hyperparameter values \\
    \midrule
         batch size & 128 $\times$ number of tasks \\
         network architecture & feedforward network\\
         actor/critic size & three fully connected layers with 400 units \\
         non-linearity & ReLU \\
         policy initialization & standard Gaussian\\
         exploration parameters & run a uniform exploration policy 1500 steps\\
         \# of samples / \# of train steps per iteration & 1 env step / 1 training step\\
         policy learning rate & 3e-4 \\
         Q function learning rate & 3e-4 \\
         optimizer & Adam\\
         policy learning rate & 3e-4 \\
         beta for Adam optimizer for policy & (0.9, 0.999) \\
         Q function learning rate & 3e-4 \\
         beta for Adam optimizer for Q function & (0.9, 0.999) \\
         discount & .99 \\
         Episode length (horizon) & 150 \\
         reward scale & 1.0 \\
    \bottomrule
    \end{tabular}
}
\end{table}

\begin{table}[ht!]
\centering
\caption{Hyperparameter values for Multi-task SAC}
\label{table::multitask_sac_hp}
\resizebox{0.5\textwidth}{!}
{
    \begin{tabular}{p{3.7cm}p{3.7cm}}
    \toprule
         Hyperparameter & Hyperparameter values \\
    \midrule
         temperature & learned and disentangled with tasks \\
    \bottomrule
    \end{tabular}
}
\end{table}

\begin{table}[ht!]
\centering
\caption{Hyperparameter values for Multi-task SAC + Task Encoder}
\label{table::multitask_sac_task_encoder_hp}
\resizebox{0.5\textwidth}{!}
{
    \begin{tabular}{p{3.7cm}p{3.7cm}}
    \toprule
         Hyperparameter & Hyperparameter values \\
    \midrule
         task encoder size & embedding layer + two fully connected layers. Embedding/hidden/output dims = 50\\
         temperature & learned and disentangled with tasks \\
    \bottomrule
    \end{tabular}
}
\end{table}

\begin{table}[t!]
\centering
\caption{Hyperparameter values for Multi-headed SAC}
\label{table::multiheaded_sac_hp}
\resizebox{0.5\textwidth}{!}
{
    \begin{tabular}{p{3.7cm}p{3.7cm}}
    \toprule
         Hyperparameter & Hyperparameter values \\
    \midrule
         network architecture & multi-head (one head per task) \\
         temperature & learned and disentangled with tasks \\
    \bottomrule
    \end{tabular}
}
\end{table}

\begin{table}[ht!]
\centering
\caption{Hyperparameter values for PCGrad}
\label{table::multitask_sac_pcgrad_hp}
\resizebox{0.5\textwidth}{!}
{
    \begin{tabular}{p{3.7cm}p{3.7cm}}
    \toprule
         Hyperparameter & Hyperparameter values \\
    \midrule
         actor/critic size & five fully connected layers with 400 units \\
         temperature & learned and disentangled with tasks \\
    \bottomrule
    \end{tabular}
}
\end{table}

\begin{table}[ht]
\centering
\caption{Hyperparameter values for Soft Modularization}
\label{table::multitask_sac_softmodularization_hp}
\resizebox{0.5\textwidth}{!}
{
    \begin{tabular}{p{3.7cm}p{3.7cm}}
    \toprule
         Hyperparameter & Hyperparameter values \\
    \midrule
         task encoder size & two layer feedforward network. Hidden/output dims = 50 \\
         routing network size & 4 layers and 4 modules per layer. \\
         temperature & learned and disentangled with tasks \\
    \bottomrule
    \end{tabular}
}
\end{table}

\begin{table}[ht!]
\centering
\caption{Hyperparameter values for FiLM}
\label{table::multiheaded_sac_film_hp}
\resizebox{0.5\textwidth}{!}
{
    \begin{tabular}{p{3.7cm}p{3.7cm}}
    \toprule
         Hyperparameter & Hyperparameter values \\
    \midrule
         task encoder size & two layer feedforward network. Hidden/output dims = 50 \\
         temperature & learned and disentangled with tasks \\
    \bottomrule
    \end{tabular}
}
\end{table}

\begin{table}[ht!]
\centering
\caption{Hyperparameter values for CARE}
\label{table::multiheaded_sac_care_hp}
\resizebox{0.5\textwidth}{!}
{
    \begin{tabular}{p{3.7cm}p{3.7cm}}
    \toprule
         Hyperparameter & Hyperparameter values \\
    \midrule
         task encoder size & two layer feedforward network. Hidden/output dims = 50 \\
         number of encoders & 6 for MT10, 10 for MT50 \\
         temperature & learned and disentangled with tasks \\
    \bottomrule
    \end{tabular}
}
\end{table}

\section{Additional Results}
\label{app::results}

\subsection{Ablations}
\label{app::results::mt10_ablations_with_encoders}
In~\cref{table::mt10_ablations_with_encoders}, we consider some ablations with the CARE model for MT10. First, we vary the number of encoders ($k$). We note that having too many encoders can hurt the performance. Second, we consider the case where we use initialise $m$ encoders but use only top-$k$ encoders at each timestep. Top-$k$ encoders are the ones which have the top-$k$ highest attention scores. The remaining $m-k$ are not used and are considered to be \textit{inactive}. The attention scores $\alpha$, corresponding to the selected $k$ encoders, are re-normalized to sum to 1. In this case, $$z_{enc} = \frac{\sum_{i}^{m}\alpha_{i} \times z_{enc}^{i} \times \mathbbm{1}_{top-k}(i)}{\sum_{i}^{m}\alpha_{i} \times \mathbbm{1}_{top-k}(i)},$$ where $\mathbbm{1}_{top-k}(i)$ indicates if the $\alpha_i$ is among the top-$k$ attention weights. We find that while this form of \textit{hard-attention} does not work as well as the \textit{soft-attention} mechanism used by CARE, the performance is still comparable to the performance of other baselines. Finally, we consider a hand-coded assignment of tasks to encoders. We read the task descriptions, identify common skills and objects, and manually assign these skills and objects to encoders. The tasks that mention these skills/objects are then mapped to the corresponding encoders. The resulting task-encoder mappings are shown in~\cref{table::mt10::clusters}. We note that this manual mapping of encoders works very well in practice and the mapping is interpretable by design. However, the process of creating such mappings is time consuming and error-prone. On the other hand, CARE can learn the mapping between tasks and encoders while learning the representations.
This experiment validates the hypothesis that explicitly assigning object and skill concepts to the encoders improves multi-task performance, and lends credibility to the additional hypothesis that CARE learns interpretable representations, i.e. ones that represent objects and skills.

\subsection{Effect of frequency of evaluation}
\label{app::results::effect_of_frequency_of_evaluation}

We find that we can improve the agent's performance just by increasing the frequency of evaluation as shown in~\cref{table::mt10::care::1Keval}. 
In Meta-World, the agent is evaluated for a binary-valued success signal and evaluating the agent more often makes it more likely for the agent to solve a given task. This effect makes it harder to compare the performance of the algorithms from different works. We control for this issue in our setup by evaluating every agent at a fixed frequency (once every 10K environment steps, per task).

\begin{table}[!ht]
\centering
\caption{Evaluation performance on the MT10 test environments, after training for \textbf{500 thousand} steps (for each environment). The results are averaged over 10 seeds. ME = Mixture of Encoders. CARE's improvement is statistically significant for the baselines marked with *.}
\label{table::mt10::care_500k}
\begin{tabular}{p{5.6cm}p{2.1cm}}
\toprule
         Agent & success \\
         & \small{(mean $\pm$ stderr)} \\
\midrule

    Multi-task SAC \cite{meta-world} * & 0.38 $\pm$ 0.041 \\
    Multi-task SAC + Task Encoder * & 0.42 $\pm$ 0.031 \\
    Multi-headed SAC \cite{meta-world} * & 0.44 $\pm$ 0.045 \\
    PCGrad \cite{gradient_surgery_for_multitask_learning} * & 0.53 $\pm$ 0.030 \\
    Soft Modularization \cite{multi_task_reinforcement_learning_with_soft_modularization} & 0.64 $\pm$  0.052 \\
    SAC + FiLM \cite{film} * &  0.57 $\pm$  0.035 \\
    SAC + Metadata + ME (\textbf{CARE}) &  \textbf{0.66} $\pm$ \textbf{0.028} \\
\bottomrule
\end{tabular}
\end{table}

\begin{table}[th!]
\centering
\caption{Evaluation performance on the MT50 test environments, after training for \textbf{500 thousand} steps (for each environment). The results are averaged over 10 seeds. The proposed method, CARE, outperforms the other baselines. ME = Mixture of Encoders. CARE's improvement is statistically significant for the baselines marked with *.}
\label{table::mt50::care_500k}
\begin{tabular}{p{5.6cm}p{2.1cm}}
\toprule
         Agent & success \\
         & \small{(mean $\pm$ stderr)} \\
\midrule

    Multi-task SAC * \cite{meta-world} & 0.28 $\pm$ 0.017 \\
    Multi-task SAC + Task Encoder * & 0.37 $\pm$ 0.016 \\
    Multi-headed SAC \cite{meta-world} & 0.45 $\pm$ 0.064 \\
    PCGrad \cite{gradient_surgery_for_multitask_learning} & 0.47 $\pm$ 0.016 \\
    Soft Modularization \cite{multi_task_reinforcement_learning_with_soft_modularization} & 0.47 $\pm$  0.012 \\
    SAC + FiLM \cite{film} * &  0.30 $\pm$  0.012\\
    SAC + Metadata + ME (\textbf{CARE}) &  \textbf{0.51} $\pm$ \textbf{0.036} \\
\bottomrule
\end{tabular}
\end{table}

\section{Testing for statistical significance}
\label{app::testing_for_statstical_significance}

We perform a two-tailed, Student's $t$-distribution test~\cite{student1908probable} under \textit{equal sample sizes, unequal variance} setup (also called Welch's $t$-test). The null hypothesis is: the mean performance of the two models (CARE and any baseline) are equal. The significance level ($p$) is set to 0.05.

\begin{table}[ht!]
\centering
\caption{Evaluation performance on the MT10 test environments, after training for \textbf{2 million} steps (for each environment). The results are averaged over 10 seeds. ME = Mixture of Encoders. CARE's improvement is statistically significant for the baselines marked with *.}
\label{table::mt10_ablations_with_encoders}
\begin{tabular}{p{5.6cm}p{2.1cm}}
\toprule
         Agent & success \\
         & \small{(mean $\pm$ stderr)} \\
\midrule
     CARE with 2 encoders &        $0.76 \pm       0.043$ \\
     CARE with 10 encoders * &        $0.71 \pm       0.029$ \\
     \midrule
     CARE with 10 encoders with 2 active * &        $0.55 \pm       0.051$ \\
     CARE with 10 encoders with 4 active * &        $0.69 \pm       0.040$ \\
     CARE with 10 encoders with 6 active * &        $0.71 \pm       0.051$ \\
     CARE with 10 encoders with 8 active * &        $0.67 \pm       0.043$ \\
     \midrule
     Manually mapping tasks to encoders &        $0.80 \pm       0.037$ \\
     \midrule
     CARE with 6 encoders &     $\mathbf{0.84} \pm       \mathbf{0.051}$ \\
\bottomrule
\end{tabular}
\vspace{-15pt}
\end{table}

\begin{table}[!bh]
\caption{Evaluation performance on the MT10 test environments, after training for \textbf{2 million} steps (for each environment). The agent is evaluated after 1K and 10K steps. The results are averaged over 10 seeds. ME = Mixture of Encoders. We note that evaluating the agent more frequently improves the performance across the baselines.}
\label{table::mt10::care::1Keval}
\begin{tabular}{p{3.5cm}p{2.0cm}p{2.0cm}}
\toprule
         Agent & success (1K) & success (10K) \\ & \small{(mean $\pm$ stderr)}  & \small{(mean $\pm$ stderr)} \\
\midrule

     PCGrad~\cite{gradient_surgery_for_multitask_learning} &         $0.75 \pm       0.033$ &        $0.72 \pm       0.022$ \\
     Soft Modularization \cite{multi_task_reinforcement_learning_with_soft_modularization} &        $0.78 \pm       0.049$ &        $0.73 \pm       0.043$ \\
     SAC + Metadata + ME (\textbf{CARE}) &     $\mathbf{0.87} \pm       \mathbf{0.054}$ &     $\mathbf{0.84} \pm       \mathbf{0.051}$ \\
\bottomrule
\end{tabular}
\end{table}

\begin{table}[th!]
\caption{Mapping between the encoders and skills/objects and tasks for MT10 environments.}
\label{table::mt10::clusters}
\begin{tabular}{p{1.0cm}p{3.0cm}p{3.5cm}}
\toprule
Encoder Index & Skills or objects that map to the encoder & Tasks that map to the encoder \\ \hline
1                      & skill: close                               & drawer-close-v1, window-close-v1                                     \\ \hline
2                      & skill: open                                & door-open-v1, drawer-open-v1, window-open-v1                         \\ \hline
3                      & skill: push                                & push-v1, door-open-v1,  drawer-open-v1, window-open-v1               \\ \hline
4                      & skill: any remaining skills                    & reach-v1, pick-place-v1, button-press-topdown-v1, peg-insert-side-v1 \\ \hline
5                      & object: drawer                             & drawer-open-v1, drawer-close-v1                                      \\ \hline
6                      & object: goal                              & reach-v1, push-v1, pick-place-v1, peg-insert-side-v1                 \\ \hline
7                      & object: puck                               & push-v1, pick-place-v1                                               \\ \hline
8                      & object: window                          & window-open-v1, window-close-v1                                      \\ \hline
9                      & object: any remaining objects                  & door-open-v1, button-press-topdown-v1, peg-insert-side-v1 \\
\bottomrule
\end{tabular}
\end{table}

\end{document}